\documentclass[conference]{IEEEtran}
\IEEEoverridecommandlockouts
\usepackage{cite}
\usepackage{amsmath,amssymb,amsfonts}
\usepackage{graphicx}
\usepackage{array}
\usepackage{textcomp}
\usepackage{xcolor}
\usepackage{amsmath}
\usepackage{algorithm}
\usepackage[noend]{algpseudocode}
\usepackage{tabularx,booktabs}
\newcolumntype{C}{>{\centering\arraybackslash}X} 
\def\BibTeX{{\rm B\kern-.05em{\sc i\kern-.025em b}\kern-.08em
    T\kern-.1667em\lower.7ex\hbox{E}\kern-.125emX}}
\newtheorem{theorem}{Theorem}

\title{\LARGE Koopcon: A new approach towards smarter and less complex learning}

\usepackage{titlesec}

\makeatletter
\newcommand{\linebreakand}{%
  \end{@IEEEauthorhalign}
  \hfill\mbox{}\par
  \mbox{}\hfill\begin{@IEEEauthorhalign}
}
\makeatother

\author{
  \IEEEauthorblockN{Vahid Jebraeeli}
  \IEEEauthorblockA{\textit{Electrical and Computer Engineering Department} \\
    \textit{North Carolina State University}\\
    Raleigh, USA \\
    vjebrae@ncsu.edu}
  \and
  \IEEEauthorblockN{Bo Jiang}
  \IEEEauthorblockA{\textit{Electrical and Computer Engineering Department} \\
    \textit{North Carolina State University}\\
    Raleigh, USA \\
    bjiang8@ncsu.edu}
  \linebreakand % <------------- \and with a line-break
  \IEEEauthorblockN{Derya Cansever}
  \IEEEauthorblockA{\textit{Electrical and Computer Engineering Department} \\
    \textit{North Carolina State University}\\
    Raleigh, USA \\
    dhcansev@ncsu.edu}
  \and
  \IEEEauthorblockN{Hamid Krim}
  \IEEEauthorblockA{\textit{Electrical and Computer Engineering Department} \\
    \textit{North Carolina State University}\\
    Raleigh, USA \\
    ahk@ncsu.edu}
}

\begin{document}
\maketitle
\begin{abstract}
    In the era of big data, the sheer volume and complexity of datasets pose significant challenges in machine learning, particularly in image processing tasks. This paper introduces an innovative Autoencoder-based Dataset Condensation Model backed by Koopman operator theory that effectively % synthesizes 
    packs large datasets into compact, information-rich representations. % Drawing inspiration from 
    Inspired by the predictive coding mechanisms of the human brain, our model leverages a novel approach to encode and reconstruct data, maintaining essential features and label distributions. The condensation process utilizes an autoencoder neural network architecture, coupled with Optimal Transport theory and Wasserstein distance, to minimize the distributional discrepancies between the original and synthesized datasets. We present a two-stage implementation strategy: first, condensing the large dataset into a smaller synthesized subset; second, evaluating the synthesized data by training a classifier and comparing its performance with a classifier trained on an equivalent subset of the original data. Our experimental results demonstrate that the classifiers trained on condensed data exhibit comparable performance to those trained on the original datasets, thus affirming the efficacy of our condensation model. This work not only contributes to the reduction of computational resources but also paves the way for efficient data handling in constrained environments, marking a significant step forward in data-efficient machine learning.\footnote{Thanks to the generous support of ARO grant W911NF-23-2-0041.}
\end{abstract}

\begin{IEEEkeywords}
Condensation, Autoencoder, Optimal Transport, Wasserstein Distance
\end{IEEEkeywords}

\section{Introduction}

% In machine learning, especially image processing, managing increasingly large and complex datasets is a paramount challenge. 
The large amounts of data required for training a Deep Learning network for computer vision or imaging problems is increasingly arising as a challenging issue. This paper introduces an innovative method, deeply rooted in the principles of predictive coding \cite{b0} observed in the human brain. Our approach aims to emulate the brain's efficiency in processing and % synthesizing 
exploiting vast quantities of sensory information. By creating compact, predictive representations from complex stimuli, our model parallels the brain's % method of accentuating 
capacity to differentially distinguish expected and observed data % differences between expected and observed data. 
This enables a significant reduction in the volume of data required for effective training and analysis. Our approach distills large-scale image datasets into condensed, information-rich synthesized counterparts, % reflecting the cognitive economy of the brain, 
adeptly balancing the richness of information with computational efficiency. % This paper details our model's development, emphasizing its alignment with neuroscientific insights, and marks a significant stride in the evolution of data-efficient learning strategies in machine learning.

% Building upon these innovative strides, 
In addressing this data condensation issue, the work of Zhao et al. in \cite{b1} % emerges as a pivotal contribution to the domain of 
was an early contribution data-efficient learning strategies in machine learning. % Their research addresses the pressing challenge of managing large volumes of data in image processing. They introduce a technique for dataset condensation, centering on the concept of gradient matching. 
% This 
Their approach % involves 
entails optimizing % synthetic 
the synthesized data so that the gradients of a corresponding neural network, when trained on these % synthetic 
synthesized datasets, to closely % mirror 
match those obtained from training on original larger datasets. The % essence of their technique lies in producing 
resulting 
condensed datasets % that 
encapsulate the core characteristics and diversity of % the original datasets effective and efficient reducing the excessive training data sizes while preserving all the necessary features to safeguard a competitive performance. neural network training without the 
extensive data requirements typically associated with such tasks. They % provide a thorough analysis 
% demonstrating
demonstrate the effectiveness of their method across various neural network architectures, % highlighting how their method creates 
with condensed datasets that are both compact and informationally dense, thus reducing computational and memory % demands.
gains.

% Furthering this line of research
In another twist on the theme, Zhao et al. \cite{b2} developed an efficient method for dataset condensation that synthesizes informative samples % by matching 
whose feature distributions % of synthetic and 
those of original training images in various sampled embedding spaces. This approach significantly reduced synthesis cost and exhibits comparable or superior performance in various settings, including larger datasets and more sophisticated neural architectures. Their method marks a notable advancement by providing a practical solution for dataset condensation % without the need for 
foregoing complex bi-level optimization, thereby easing the computational burden in % machine learning 
training processes.

% Additionally, the work of 
Cazenavette et al. in \cite{b3} % presents another significant contribution. They 
introduce a % unique 
technique for dataset condensation % through the imitation of 
using the notion of long-range training dynamics % . The core innovation lies in using
by exploiting "expert trajectories" of sequences of model parameters obtained during training on a full dataset. By matching segments of these expert trajectories with those derived from models trained on % synthetic
synthesized data, their approach effectively captures the essential learning dynamics necessary for training % neural networks. 
the latter. % This method not only outperforms existing dataset distillation techniques but also demonstrates the potential for scaling to higher-resolution data, traditionally a challenge in the field.
This proposed approach achieved better performance and a promising scaling potential to higher resolution data.

% Our paper contributes to this dynamic field by introducing 
We  introduce here an Autoencoder-based % Dataset 
Model % , synthesizing 

for condensing large datasets into compact, information-rich representations. % Our approach, drawing inspiration from 
Inspired by predictive coding \cite{b0} % mechanisms of the human brain
which conjectured to be central to brain cognitive functionality, we employ a combination of Optimal Transport \cite{b9} theory and Wasserstein distance \cite{b8} to minimize discrepancies between the original and synthesized datasets. % The efficacy of our model is validated through extensive experiments, demonstrating that classifiers trained on condensed data perform comparably to those trained on original datasets. This significantly reduces computational resources and facilitates efficient data handling in constrained environments.
We demonstrate the condensation viability of this novel  approach, by conceptually projecting the nonlinear data features onto a space with a learned so-called Koopman operator, and subsequently exploiting the latter's properties for a systematic realization. 

The paper's structure is as follows: Section II provides background on Koopman Operator Theory \cite{b4} and its application in deep learning \cite{b5}. Section III introduces the Koopcon model, explaining its formulation, design, and how it % incorporates
integrates elements like autoencoder architecture, self-attention mechanisms \cite{b10}, and optimal transport theory \cite{b9}. Section IV presents our experimental approach, demonstrating the model's effectiveness in maintaining classifier performance with condensed datasets and comparing it against existing methods on standard datasets. The final section, V, concludes the paper, summarizing key findings and outlining potential future directions.

% In summary, our work sets a new standard in dataset efficiency for machine learning, effectively preserving the integrity and label distribution of the original datasets while ensuring robust classifier performance across a range of architectures.

\section{Related Background}

\subsection{Koopman Operator Theory}

Koopman operator theory offers a rich and elegant framework for analyzing nonlinear dynamical systems by transforming them into a linear context. % The theory, first proposed by Bernard Koopman in 1931, 
First purposed by Koopman in \cite{b4}, the theory facilitates the study of complex systems using linear operators on function spaces, regardless of the nonlinearity in the state space.

\begin{theorem}[Koopman Operator Linearity]
Given a nonlinear dynamical system with state evolution defined by $\vec{x}_{t+1} = f(\vec{x}_t)$, where $\vec{x}_t$ (the system state at time $t$) $\in \mathcal{M} \subseteq \mathbb{R}^n$ and $f: \mathcal{M} \rightarrow \mathcal{M}$, the Koopman operator $\mathcal{K}: \mathcal{H} \rightarrow \mathcal{H}$ acts linearly on observable functions $g: \mathcal{M} \rightarrow \mathbb{R}$ in the Hilbert space $\mathcal{H}$, such that:
\begin{equation}
    (\mathcal{K}g)(x_t) = g(f(x_t)) = g(x_{t+1})
\end{equation}
\end{theorem}

The theorem emphasizes that the Koopman operator advances observables $g$ linearly in time according to the system's dynamics. The eigenfunctions $\psi$ of the Koopman operator satisfy the linear eigenvalue equation $\mathcal{K}\psi(x) = \lambda \psi(x)$, with $\lambda$ as the eigenvalue, indicating a scaled or rotated evolution of the eigenfunction, % but 
with its structure preserved over time.

\subsection{Deep Koopman Operator}
Expanding upon Koopman's theory, recent advancements in deep learning \cite{b5} have facilitated the approximation of the Koopman operator using neural networks, allowing for practical applications in a variety of complex systems.

\begin{theorem}[Deep Koopman Learning]
Observations $X = [x_1, x_2, \dots, x_t]$ and their time-evolved states $X' = [x_2, x_3, \dots, x_{t+1}]$ of a dynamical system can be utilized to learn a neural network approximation of the Koopman eigenfunctions $\phi(\cdot)$ and the linear dynamics embodied by a matrix $T$ in state space $\mathcal{Y}$, by minimizing the loss function:
\begin{equation}
    \min_{\phi, T} \|\hat{X}' - X'\|^2,
\end{equation}
where $\hat{X}'$ are the predicted future states generated by the learned dynamics.
\end{theorem}

% This deep learning approach employs an autoencoder architecture to simultaneously learn the encoding function $\phi(\cdot)$, which captures the Koopman eigenfunctions, and the linear dynamics matrix $T$. 
We adopt an autoencoder to capture the K-eigenfunction $\phi(\cdot)$ and the linear dynamics matrix $T$. The encoder maps the input data into a latent space representing the Koopman observables, and the decoder reconstructs the state space from these observables. The linear evolution in the latent space is governed by the learned matrix $T$, analogous to the Koopman matrix $K$, facilitating the prediction of future system states. % The integration of Koopman theory with deep learning, as detailed in "DLKoopman: A Deep Learning Software Package for Koopman Theory"
As described in \cite{b5}, DLKoopman % represents a significant leap in the study of dynamical systems. It 
provides a bridge between non-linear dynamics and linear predictive models. % , allowing for the application of linear systems theory to a broad class of problems that were previously intractable using conventional methods.

\begin{figure}
  \centering
  \includegraphics[width=8.5cm]{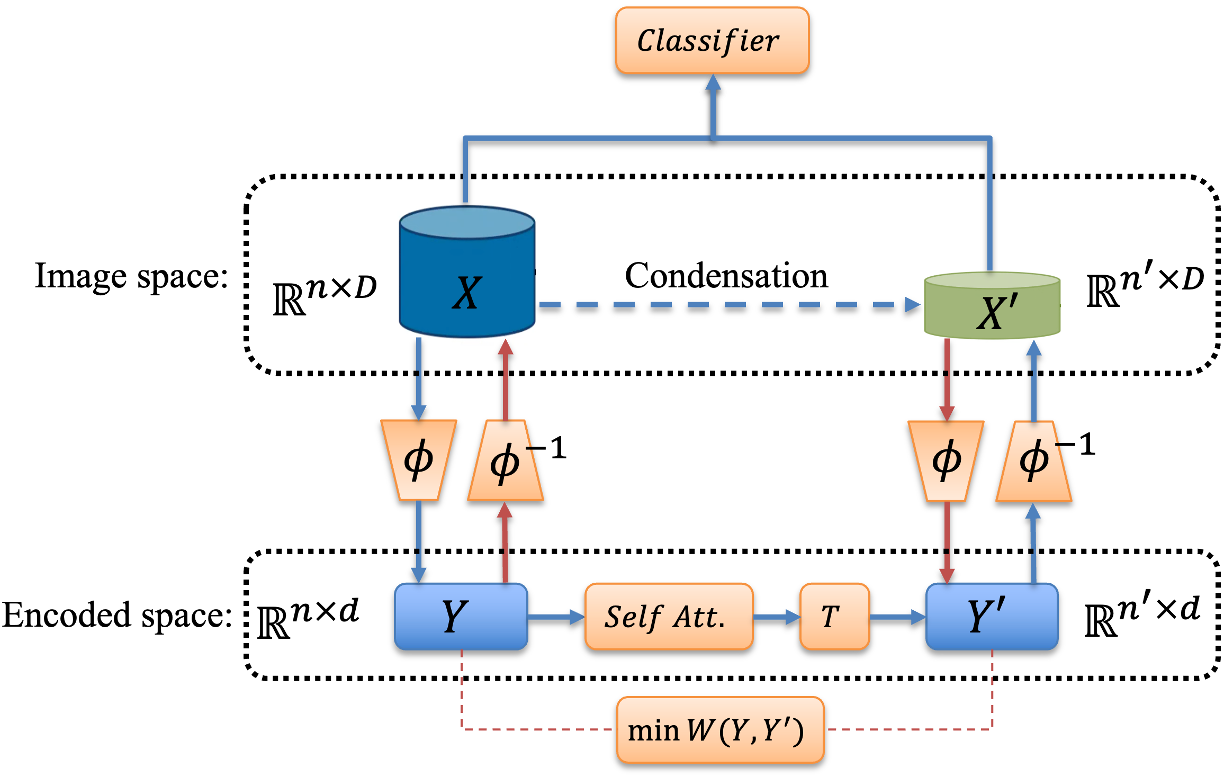}
  \caption{Our proposed model architecture}
  \label{fig:DeepKoopmanTheory}
\end{figure}

% \begin{figure}
%   \centering
%   \includegraphics[width=8.5cm]{self_att.png}
%   \caption{Self-Attention \cite{b9} model used in our condensation model}
%   \label{fig:DeepKoopmanTheory}
% \end{figure}

\subsection{Notations}

% To ensure 
For clarity and consistency throughout this paper, the following notations are adopted, where capital  boldface lower case letters respecttively denote matrices and vectors, and subscripts denote vector elements while superscripts indicate an alternative copy in a sequence or a transformation.

\begin{itemize}
  \item \( X \): The original high-dimensional and large-scale dataset, where each element \( x_i \) represents an individual data point with associated features.
  
  \item \( Y \): The latent representation of \( X \) obtained after encoding through the autoencoder's encoder network \( \phi \).
  
  \item \( X' \): The condensed dataset synthesized from \( X \), which is smaller in size but designed to retain the essential information of the original dataset.
  
  \item \( Y' \): The condensed latent representation of \( Y \), which is the result of applying the condensation process within the latent space.
  
  \item \( \phi \): The encoder part of the autoencoder that maps the input data \( X \) to its latent representation \( Y \).
  
  \item \( \phi^{-1} \): The decoder part of the autoencoder that maps the latent representation \( Y \) back to the reconstructed data \( \hat{X} \) or \( X' \).

  \item \( f_c \): The classifier function trained on the reconstructed condensed dataset \( X' \).
  
  \item \( \mathcal{W} \): The Wasserstein distance used to measure distributional discrepancies in the condensation process.
\end{itemize}

% Please note that vectors are denoted in boldface (e.g., \( \boldsymbol{x} \)), matrices in uppercase (e.g., \( Q \)), and scalar quantities in regular typeface (e.g., \( n \)). Subscripts are used to refer to elements of vectors or matrices, and superscripts indicate different versions or transformations of the data.

\subsection{Optimal Transport}

Data condensation may be abstractly interpreted as a generalized compression/dimension reduction in that a number of data entities (e.g., images) are compressed into one entity, while methodically aggregating all associated characteristic features. In so doing, we seek to quantitatively track this task by comparing the resulting distributions to that prior, using a measure derived from optimal transport theory. This approach draws upon recent advancements (like in \cite{b11} and \cite{b12}), integrating principles of optimal transport to enhance the fidelity and efficiency of data condensation, thereby preserving essential information while achieving significant reductions in data volume.

Integral to our model is the minimization of the cost \(c\) for transforming the encoded latent representation of original data with a probability density function \(p(Y)\) to closely match that of condensed version of data \(p(Y')\). This is articulated through the Optimal Transport Loss:
\begin{equation}
    \mathcal{L}_{O.T.}=\min _{\pi \in \Pi(p(Y), p(Y'))} \mathbb{E}_{(Y, Y') \sim \pi}[c(Y, Y')]
\end{equation}
Here, \(\pi\) represents a coupling between the distributions \(p(Y)\) and \(p(Y')\), with \(c(Y, Y')\) denoting the dissimilarity measure between \(Y\) and \(Y'\).

\subsection{Problem Formulation}

The dataset condensation problem involves transforming a large-scale training set \( X \) into a smaller synthetic set \( X' \). Formally, \(X = \left\{ \left(\boldsymbol{x}_1, y_1\right), \ldots, \left(\boldsymbol{x}_n, y_n\right) \right\}\) comprises \( n \) image and label pairs. The condensed set \( X' = \left\{ \left(\boldsymbol{x'}_1, y_1\right), \ldots, \left(\boldsymbol{x'}_{n'}, y_{n'}\right) \right\} \) % contains 
yields \( n' \) synthetic image and label pairs. The % fundamental 
principle objective is to % ensure that models trained on both \( X \) and \( X' \) achieve comparable performance on unseen testing data, formulated as:
seek CNN-parameterized models of \( X \) and \( X' \) which similarly perform on unseen testing data, analytically stated as:
$$
\mathbb{E}_{\boldsymbol{x} \sim P_{\mathcal{X}}}\left[\ell\left(\Psi_{\boldsymbol{\theta}^{X}}(\boldsymbol{x}), y\right)\right] \simeq \mathbb{E}_{\boldsymbol{x} \sim P_{\mathcal{X}}}\left[\ell\left(\Psi_{\boldsymbol{\theta}^{X'}}(\boldsymbol{x}), y\right)\right],
$$
where \( P_{\mathcal{X}} \) represents the real data distribution, \( \ell \) the loss function (e.g., cross-entropy), and \( \Psi \) a deep neural network parameterized by \( \boldsymbol{\theta} \).

\section{Methodology}

% Our Koopcon model,
The driving intuition behind  our proposed  solution to data Koopman-condensation (Koopcon) lies in the projection of  characteristic features \( Y \) in data at hand, and in their systematic,  class-consistent and distributionally-balanced packing \( Y' \)  prior to reconstruction (\( X' \)). 

{\bf Claim:} {\bf \it A near-optimal and inference-driven dimension-reduction $Y'$ of a data-set $X$, can be achieved by an optimal transport in a Koopman-data space.}\\
As depicted in Figure 1, is engineered to condense a voluminous, high-dimensional dataset \( X \) into the compact yet informatively dense set \( X' \) % . This synthesized set 
is crafted to retain the crucial attributes of the original dataset exploiting the Koopman linear evolution of non-linear dynamics.
% . The model is rooted in Koopman Theory, which posits that even complex, nonlinear dynamics can be linearized within an expanded linear framework, thereby simplifying their analysis and understanding.

The latent representation $Y \in \mathbb{R}^{n\times d}$ is the result of $\phi:\mathbb{R}^{n\times D} \rightarrow \mathbb{R}^{n \times d}$ with $d<D$, and is followed by a linear transformation to capture the intrinsic data dynamics.
% The dataset \( X \in \mathbb{R}^{n \times D} \), undergoes an initial transformation via the encoder \( \phi \), functioning as a nonlinear mapping. % This encoder compresses \( X \) into a reduced latent space, represented as \( Y \in \mathbb{R}^{n \times d} \), where \( d \) signifies a lower dimensionality that encapsulates the intrinsic dynamics inherent in \( X \).

\textbf{Reconstruction Loss (\(\mathcal{L}_{re}\)):} % This is the loss incurred when the original data \(X\) is encoded to \(Y\) and then decoded back to \(\hat{X}'\), measured against the original input \(X\):
The auto-encoder parametrization denoted by $\phi$ and $\theta$ for the encoder and decoder respectively follows the standard  optimization loss  between the input distribution and that of the output written as

\begin{equation}
    \begin{aligned}
    \mathcal{L}_{re}(\phi, \theta ; X)= 
    &\mathbb{E}_{Y \sim q_\phi(Y \mid X)}[-\log p_\theta(X \mid Y)] \\
    &+ \text{KL}[q_\phi(Y \mid X) || p(Y)],
    \end{aligned}
\end{equation}

where the first term is the expected negative log-likelihood, and the second term is the Kullback-Leibler divergence between the encoded distribution \(q_\phi(Y | X)\) and a prior distribution \(p(Y)\). % Note that a similar loss labeled $"T"$ with corresponding input/output ($Y'/X'$) is used on the reconstruction stage of the condenser.

To selectively exploit the most relevant features and further refine the intrinsic linear evolution of the nonlinear dynamics, we precede the linear transformation $T(\cdot)$ with a self-attention transformation on $Y$ to yield $Y'\in \mathbb{R}^{n'\times d}$.

% In this condensed latent space, we employ a self-attention mechanism \cite{b9} to further refine \( Y \). This self-attention is strategically selected for its proficiency in isolating and amplifying pivotal features within \( Y \), thus effectively setting the stage for the subsequent linear transformation. % The application of this self-attention prepares \( Y \) for a transformation that aligns with Koopman Theory's linear perspective. The self-attentive transformation \( T(Y) \), when subjected to a linear operation \( T \), yields a distilled latent representation \( Y' \in \mathbb{R}^{n' \times d} \). It is this representation \( Y' \) that aims to distill the core dynamics of the initial dataset \( X \) while maintaining a compact form.

% The incorporation of the self-attention mechanism is a deliberate choice, motivated by its proven effectiveness in pattern recognition and neural processing tasks. By selectively enhancing relevant features and diminishing less significant ones, self-attention ensures that our model captures a rich, feature-focused representation of the dataset, which is vital for the fidelity of the condensed data and its utility in downstream machine learning applications.

In the condensation phase, \( Y' \) crafted to % contain fewer data points (\( n' \)) 
be of lower dimension (\( n' \)) while still reflecting the original dataset's distribution. This process is guided by the minimization of the Wasserstein distance \( \mathcal{W}(Y, Y') \), ensuring that the condensed data \( Y' \) maintains the distributional integrity of \( Y \).

\textbf{Wasserstein Distance (\(\mathcal{L}_{\mathcal{W}}\)):} The concept of Wasserstein distance \cite{b8} arises as a specialized form of the Optimal Transport Loss \cite{b9}, where it specifically measures the cost to align the distribution of the encoded data \(Y\) with that of the condensed representation \(Y'\). The Wasserstein distance, therefore, quantifies the minimal "effort" required to morph the distribution \( p_Y \) into \( p_{Y'} \), making it a natural measure for the effectiveness of dataset condensation processes.

The Wasserstein distance can be expressed as:
\begin{equation}
    \begin{aligned}
    \mathcal{L}_{\mathcal{W}}(p_{Y}, p_{Y'}) = \min_{\pi \in \Pi(p_{Y}, p_{Y'})} \iint c(Y, Y') \pi(p_{Y}, p_{Y'}) \, dY \, dY'
    \end{aligned}
\end{equation}

In this formulation, \( \pi \) corresponds to the optimal transport plan that associates the distributions \( p(Y) \) and \( p(Y') \). By minimizing the Wasserstein distance $\mathcal{L}_{\mathcal{W}}$, we aim to ensure that the condensed dataset \( Y' \) not only statistically resembles the original dataset \( Y \) but also preserves its geometric and topological properties, crucial for maintaining the fidelity of the condensed data for subsequent learning tasks that depend on the intricate relationships within the data's manifold structure.

The condensed representation \( Y' \) is subsequently mapped back into the high-dimensional image space using the same decoder function \( \phi^{-1} \) that was initially used for encoding. This results in the condensed dataset \( X' \), where \( X' \in \mathbb{R}^{n' \times D} \). The utilization of the same autoencoder for both encoding and decoding stages ensures that the condensed data \( X' \) is a plausible output of the autoencoder, retaining the structure and distributional properties of the original dataset.

A classifier \( f_c \) is then trained on the reconstructed condensed dataset \( X' \), which is equipped to predict the output labels \( \hat{y} \) as if it were trained on the original dataset \( X \). This process allows the classifier to benefit from the distilled information within \( X' \), enabling efficient training with significantly reduced computational resources.

\textbf{Classification Loss (\(\mathcal{L}_{ce}\)):} Central to our model is the classification loss, which serves as a form of implicit feedback information. It evaluates the discrepancy between the predicted labels obtained from the classifier and the true labels, guiding the latent representation towards maintaining label consistency. Crucially, this process involves both the original and synthesized condensed images (\(X\) and \(X'\)), which are merged and passed through the classifier to ensure comprehensive learning. The Cross-Entropy (CE) loss metric is utilized for this purpose:
\begin{equation}
    \mathcal{L}_{ce}(f_c, \tilde{X}, y) = -\sum_{i} y_i \log(f_c(\tilde{X}_i))
\end{equation}

Here, \(f_c\) represents the classifier function, \(\tilde{X}\) is the combined set of original and reconstructed data, \(y\) is the vector of true labels, and \(\tilde{X}_i\) refers to the \(i\)-th data instance in the merged dataset. This loss component is instrumental in ensuring that the condensed dataset encapsulates not only the structural attributes of the original data but also its label characteristics, thus preserving essential discriminative features and preventing the dilution of categorical information during the dataset condensation process.

The Koopcon model leverages the computational efficiency of linear dynamics in the encoded space and the cognitive economy of the brain's predictive coding strategy \cite{b0}. It presents a significant advancement in creating data-efficient learning strategies, allowing for scalable training on extensive datasets while maintaining performance parity with models trained on the full dataset.

\textbf{Covariance Loss (\( \mathcal{L}_{cov} \)):} To foster a more diverse and representative condensed dataset, we introduce a covariance loss term into the overall loss function. This term serves as a regularizer, promoting the capture of distinct features within the latent representations \( Y \). When examining the encoded versions of all \( Y_i \)s against the encoded version of representative \( Y'_i \)s, the necessity for such a loss term becomes apparent. In scenarios without the covariance loss, the representatives tend to cluster together, leading to a less diversified representation. Conversely, the inclusion of covariance loss encourages a more scattered distribution of representatives \( Y'_i \)s, thereby enhancing the diversity within the dataset.

The mathematical definition of Covariance Loss is given by:

\begin{equation}
    \mathcal{L}_{cov}(Y') = \left\| \text{Cov}(Y') - I \right\|_F^2,
\end{equation}

where \( \text{Cov}(Y') \) denotes the covariance matrix of the latent representation \( Y' \), \( I \) is the identity matrix, and \( \left\| \cdot \right\|_F \) represents the Frobenius norm. By minimizing \( \mathcal{L}_{cov} \), the model is encouraged to produce features that are uncorrelated, thereby increasing the informativeness and variability of the synthesized samples. This discourages feature redundancy, which is instrumental in avoiding overfitting and improving the model's ability to generalize from synthesized representatives to unseen data. Thus, the Covariance Loss plays a pivotal role in ensuring that the condensed dataset is not only a compressed version of the original data but also a functionally diverse subset that retains the original's rich feature set.

% \begin{figure}
%   \centering
%   \includegraphics[width=5cm]{Diversity.png}
%   \caption{Encoded data points \( Y_i \)s (blue) and their diversified representatives \( Y'_i \)s (red) after applying covariance loss.}
%   \label{fig:DeepKoopmanTheory}
% \end{figure}

This yields a weighted sum of these losses  including a covariate spread constraint $\mathcal{L}_{cov}$ to ensure that the synthesized samples are diverse and  persistent for a good cover of the distribution, as:
\begin{equation}
   \mathcal{L}_{\text{total}} = \alpha_0 \mathcal{L}_{re} + \alpha_1 \mathcal{L}_{ce} + \alpha_2 \mathcal{L}_{\mathcal{W}} + \alpha_3 \mathcal{L}_{cov}
\end{equation}
% where \( \mathcal{L}_{ce} \) is the classification loss, \( \mathcal{L}_{cov} \) represents a covariance loss, and
where \( \alpha_0, \alpha_1, \alpha_2, \) and \( \alpha_3 \) are hyperparameters that balance the different components of the loss function.

In summary, the model synthesizes a dataset \(X'\) that, when used to train a machine learning model, aims to achieve performance comparable to using the original dataset \(X\). The losses guide the model to learn a latent representation that is both accurate to the original data and informative for the learning task. Find detailed algorithms for train and test of our model in algorithms 1 and 2 respectively.
% as detailed in Algortihmic Statement in Tables A and AA.

\begin{algorithm}
\caption{Koopcon Training Algorithm}
\begin{algorithmic}[1]
\State \textbf{Given:}
\State $X \in \mathbb{R}^{n \times D}$, original dataset with $n$ samples
\State $\phi$: Encoder mapping $\mathbb{R}^D \to \mathbb{R}^d$
\State $\phi^{-1}$: Decoder mapping $\mathbb{R}^d \to \mathbb{R}^D$
\State $n'$: Target number of synthesized samples
\State $\alpha_0, \alpha_1, \alpha_2, \alpha_3$: Weights for loss components
\State $\text{N}$: Number of training epochs
\State $\text{M}$: Number of classes of data
\State \textbf{Initialize:} Parameters of Autoencoder ($\phi$, $\phi^{-1}$), Classifier $f_c$, Linear transformation T, Self attention $SA$

\For{$\text{epoch} = 1$ \textbf{to} $\text{N}$}
    \For{$\text{class} = 1$ \textbf{to} $\text{M}$}
        \State $Y \gets \phi(X)$
        \State $Y' \gets \text{T}(SA(Y))$
        \State $X' \gets \phi^{-1}(Y')$
        \State $L_{re} \gets ||X' - X||^2$
        \State $\hat{Y} \gets f_c(X \oplus X'), (\oplus \text{: concatenation})$  
        \State $L_{ce} \gets -\sum_i y_i \cdot \log(\hat{Y}), (y_i \text{: vector of true labels})$
        \State $\mathcal{L}_{\mathcal{W}} \gets \mathcal{W}(Y, Y'), (\mathcal{W} \text{: Wasserstein Distance})$
        \State $\mathcal{L}_{cov}(Y') \gets \left\| \text{Cov}(Y') - I \right\|_F^2$
        \State $L_{total} \gets \alpha_0 L_{re} + \alpha_1 L_{ce} + \alpha_2 \mathcal{L}_{\mathcal{W}} + \alpha_3 \mathcal{L}_{cov}$
        \State \text{Update Parameters}
    \EndFor
\EndFor
\end{algorithmic}
\end{algorithm}

\begin{algorithm}
\caption{Koopcon Testing Algorithm}
\begin{algorithmic}[1]
\State \textbf{Given:}
\State $X_{\text{train}}, X_{\text{test}}$, real train and test data
\State $\phi$: Encoder mapping $\mathbb{R}^D \to \mathbb{R}^d$
\State $\phi^{-1}$: Decoder mapping $\mathbb{R}^d \to \mathbb{R}^D$
\State Linear transformation $T$ and Self-Attention $SA$
\State Classifiers $f_c$
\State $\text{N}$: Number of training epochs
\State $\text{M}$: Number of classes
\State \textbf{Initialize:} Load trained parameters for Autoencoder ($\phi$, $\phi^{-1}$), Linear Transformation $T$, and Self-Attention $SA$

\For{$\text{epoch} = 1$ \textbf{to} $\text{N}$}
    \For{$\text{class} = 1$ \textbf{to} $\text{M}$}
        \State $Y \gets \phi(X_{\text{train}})$
        \State $Y' \gets T(SA(Y))$
        \State $X' \gets \phi^{-1}(Y')$
        \State $f_c^{\text{synth}}$: Train classifier $f_c$ on $X'$
        \State $f_c^{\text{real}}$: Train classifier $f_c$ on $X_{\text{train}}$
    \EndFor
\EndFor

\State \textbf{Evaluate:} Test $f_c^{\text{synth}}$ and $f_c^{\text{real}}$ on $X_{\text{test}}$
\State \textbf{Compare Performance:} Calculate and report accuracy for both classifiers
\end{algorithmic}
\end{algorithm}

\section{Experiments and Results}

\subsection{Stages of Implementation}

The stages of implementation are illustrated in Figure 2, which outlines the two-phase process of dataset condensation and subsequent evaluation.

\subsubsection{First Stage (Condensation)}  

\textbf{I. Input (Real Big Dataset):} We begin with a large dataset \( X \), consisting of pairs \( (x_i, y_i) \) where \( x_i \) represents the features (e.g., images) and \( y_i \) the corresponding labels. The dataset has \( n \) such pairs, and labels range over \( C \) different classes, from 0 to \( C-1 \).

\textbf{II. Dataset Condensation Process:} This large dataset \( X \) undergoes a condensation process to produce a much smaller, synthesized dataset \( X' \). This condensed dataset contains pairs \( (x'_i, y_i) \), where \( x'_i \) are the synthesized features (condensed representations) and \( y_i \) are the corresponding labels. There are \( n' \) pairs in \( X' \), and it maintains the same range of labels as the original dataset.

\subsubsection{Second Stage (Evaluation)}   

\textbf{I. Training with Synthetic Data (Small Dataset):} The synthesized dataset \( X' \) is then used to train a classifier. The classifier learns to predict labels based on the condensed feature set provided by \( X' \).

\textbf{II. Training with Real Data (Big Dataset):} In parallel, you train the same type of classifier on a subset of the original large dataset \( X \). This subset is selected to have the same number of examples \( n' \) as the synthesized dataset to make a fair comparison.

\textbf{III. Comparison of Test Performance:} After both classifiers are trained, their performance is evaluated on a test set. The goal is to demonstrate that the classifier trained on the synthesized dataset \( X' \) performs similarly to the classifier trained on the real dataset \( X \), despite \( X' \) being significantly smaller in size.

The underlying hypothesis is that if the synthesized dataset \( X' \) is a good condensation of \( X \), then the classifier trained on \( X' \) should generalize almost as well as the classifier trained on \( X \) when evaluated on unseen data. This would show that \( X' \) successfully captures the core information from the larger dataset \( X \), enabling effective training with much less data.

\begin{figure}
  \centering
  \includegraphics[width=8.5cm]{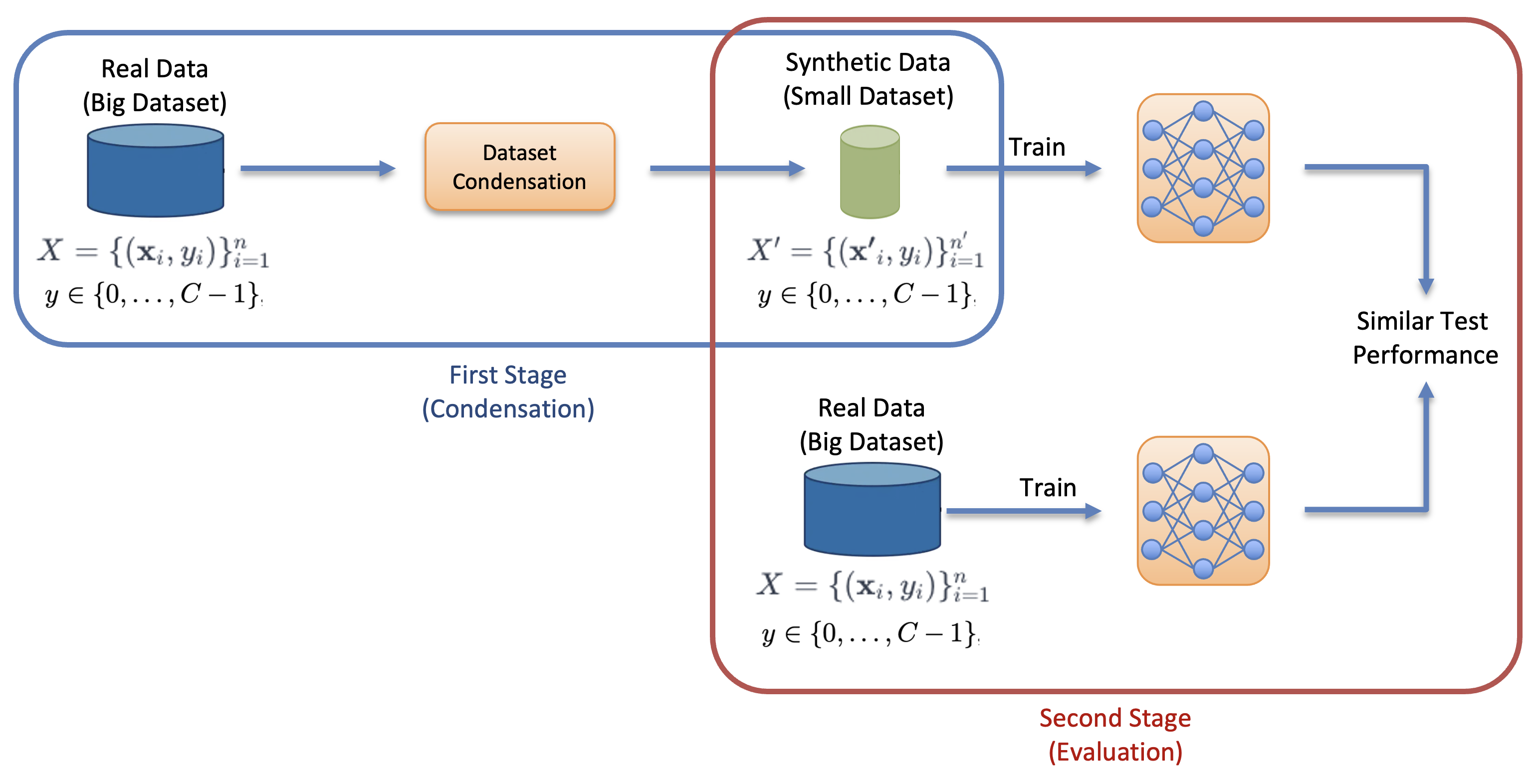}
  \caption{Stages of Implementation and evaluation of a condensation model}
  \label{fig:DeepKoopmanTheory}
\end{figure}

\subsection{Results}
The tables provide a comparative overview of classification accuracies achieved by various dataset condensation models across standard datasets CIFAR10 \cite{b13}, FashionMNIST \cite{b14}, and MNIST \cite{b15}. Each model's performance is evaluated based on the number of images per class used during training, ranging from a single image to fifty images. The reported accuracies are the mean values derived from 10 separate experiments, with the upper and lower accuracy bounds presented within the tables for each dataset.

\textbf{Table I} outlines the accuracy on the MNIST dataset. It compares our proposed model against several other models, including Gradient Matching (GM) \cite{b1}, Distribution Matching (DM) \cite{b2}, Matching Training Trajectory (MTT) \cite{b3}, Kernel Inducing Points (KIP) \cite{b6}, and the results from using the entire dataset for training. The accuracy of the proposed model consistently increases with the number of images per class, suggesting that it can effectively utilize additional data. Notably, the proposed model outperforms the other methods in each category, inching closer to the full dataset's performance as the number of images increases.

\textbf{Table II} shows the accuracies on the FashionMNIST and CIFAR10 datasets. It follows a similar pattern to Table I, where the proposed model generally surpasses the competing methods across varying images per class. The proposed model's performance on both FashionMNIST and CIFAR10 datasets shows a significant improvement over other methods, especially as the number of images per class grows.

\begin{table}[htbp]
\caption{Classification Accuracy on MNIST Dataset}
\label{my-label}
\resizebox{8.8cm}{!}{%
\begin{tabularx}{\columnwidth}{@{} c *{6}{C} @{}}
\toprule
Img/Cls & GM \cite{b1} & DM \cite{b2} & MTT \cite{b3} & KIP \cite{b6} & Ours & Whole Dataset \\ 
\midrule
1  & 91.7$\pm$0.5 & 89.7$\pm$0.6 & 91.4$\pm$0.9 & 90.1$\pm$0.1 & \textbf{95.5}$\pm$\textbf{0.5} &            \\
10 & 97.4$\pm$0.2 & 97.5$\pm$0.1 & 97.3$\pm$0.1 & 97.5$\pm$0.0 & \textbf{98.2}$\pm$\textbf{0.1} & 99.6$\pm$0.0 \\
50 & 98.8$\pm$0.2 & 98.6$\pm$0.1 & 98.5$\pm$0.1 & 98.3$\pm$0.1 & \textbf{99.4}$\pm$\textbf{0.0} &            \\
\bottomrule
\end{tabularx}
}
\end{table}

\begin{table*}[!htbp]
\caption{Classification Accuracy on FashionMNIST and CIFAR10 Datasets}
\centering
\resizebox{17.8cm}{!}{%
\begin{tabular}{ccccccccccccc}
\toprule
\multicolumn{1}{l}{} & \multicolumn{6}{c}{FashionMNIST}           & \multicolumn{6}{c}{CIFAR10}                \\ 
\cmidrule(lr){2-7} \cmidrule(lr){8-13}
Img/Cls & GM \cite{b1} & DM \cite{b2} & MTT \cite{b3} & KIP \cite{b6} & Ours & Whole Dataset & GM \cite{b1} & DM \cite{b2} & MTT \cite{b3} & KIP \cite{b6} & Ours & Whole Dataset \\
\midrule
1  & 70.5$\pm$0.6 & 71.5$\pm$0.5 & 75.3$\pm$0.9 & 73.5$\pm$0.5 & \textbf{76.0}$\pm$\textbf{0.9} &             & 28.3$\pm$0.5 & 26.0$\pm$0.8 & 46.3$\pm$0.8 & 49.9$\pm$0.2 & \textbf{51.4}$\pm$\textbf{0.7} &            \\
10 & 82.3$\pm$0.4 & 83.6$\pm$0.2 & 87.2$\pm$0.3 & 86.8$\pm$0.1 & \textbf{87.5}$\pm$\textbf{0.4} & 93.5$\pm$0.1 & 44.9$\pm$0.5 & 48.9$\pm$0.6 & 65.3$\pm$0.7 & 62.7$\pm$0.3 & \textbf{67.7}$\pm$\textbf{0.5} & 84.8$\pm$0.1 \\
50 & 83.6$\pm$0.4 & 88.2$\pm$0.1 & 88.3$\pm$0.1 & 88.0$\pm$0.1 & \textbf{88.9}$\pm$\textbf{0.0} &             & 53.9$\pm$0.5 & 63.0$\pm$0.4 & 71.6$\pm$0.2 & 68.6$\pm$0.2 & \textbf{73.2}$\pm$\textbf{0.3} &            \\
\bottomrule
\end{tabular}%
}
\end{table*}

\textbf{Table III} showcases a comparison of generalizability for various dataset condensation models by evaluating their performance across different neural network architectures. The classification accuracy percentages are reported for each condensation model when used to train four distinct classifying networks: ConvNet \cite{b16}, ResNet-18 \cite{b17}, VGG-11 \cite{b18}, and AlexNet \cite{b19}.

The comparison reveals how well each condensation approach can adapt to different architectures, which is indicative of its ability to capture the essential features of the original dataset and generalize from it. Notably, the proposed model demonstrates competitive accuracy across all classifying networks, suggesting that it produces a synthetic dataset that effectively generalizes and maintains the integrity of the data's underlying structure, regardless of the classifier used. This trait is particularly valuable in machine learning, where the ability to perform well across various architectures is a hallmark of a robust condensation technique.

\begin{table}[!htbp]
\caption{Comparative Generalization Performance of Condensation Models Across Different Classifiers}
\centering
\resizebox{8.8cm}{!}{%
\begin{tabular}{ccccc}
\toprule
\multicolumn{1}{l}{}      & \multicolumn{4}{c}{Classifying Network}                \\ 
\cmidrule(lr){2-5}
Condensation Architecture & ConvNet \cite{b16} & ResNet \cite{b17} & VGG \cite{b18}  & AlexNet \cite{b19} \\
\midrule
GM \cite{b1}    & 53.2$\pm$0.8                   & 42.1$\pm$0.7                   & 46.3$\pm$1.3                   & 34.0$\pm$2.3                   \\
DM \cite{b2}    & 49.2$\pm$0.8                   & 36.8$\pm$1.2                   & 41.2$\pm$1.8                   & 34.9$\pm$1.1                   \\
MTT \cite{b3}   & \textbf{64.4}$\pm$\textbf{0.9} & 49.2$\pm$1.1                   & 46.6$\pm$2.0                   & 34.2$\pm$2.6                   \\
KIP \cite{b6}   & 62.7$\pm$0.3                   & 49.0$\pm$1.2                   & 30.1$\pm$1.5                   & 57.2$\pm$0.4                   \\
Ours            & \textbf{65.0}$\pm$\textbf{0.5} & \textbf{60.7}$\pm$\textbf{1.0} & \textbf{59.5}$\pm$\textbf{1.5} & \textbf{61.5}$\pm$\textbf{0.9} \\
\bottomrule
\end{tabular}%
}
\end{table}

The structural variations of autoencoder architectures employed in our study include Shallow, Medium, and Deep. Each architecture represents a different level of model complexity. The Shallow autoencoder consists of 5 convolutional layers, ascending to the Medium autoencoder with 7 convolutional layers, and peaking with the Deep architecture, which comprises 9 convolutional layers. These designs are purposefully crafted to evaluate the impact of depth on the model's ability to encode and reconstruct image data, with the hypothesis that deeper networks may capture more abstract features but could also be prone to overfitting.

% \begin{figure}   
%   \centering
%   \includegraphics[width=8.5cm]{depth.png}
%   \caption{(A) Shallow (B) Medium and (C) Deep Autoencoder Architectures}
%   \label{fig:DeepKoopmanTheory}
% \end{figure}

Figure 3 presents the empirical results of our experiments, graphing the test accuracy achieved by each autoencoder architecture on the different dataset. The x-axis plots the number of images per class (Img/Cls) used during training, serving as a measure of dataset size and richness. The y-axis quantifies the test accuracy, providing a clear performance metric for each model. The lines for each architecture variant—Shallow, Medium, and Deep—converge towards the accuracy obtained when the entire dataset is leveraged for training, depicted by the 'Whole Dataset' line. These results are instrumental in understanding how the depth of an autoencoder affects its capacity for dataset condensation and subsequent classification performance, with the aim to optimize the trade-off between model complexity and generalizability.

\begin{figure}
  \centering
  \includegraphics[width=8.5cm]{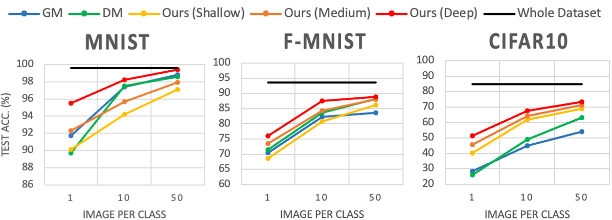}
  \caption{Test accuracy comparison between our model with different depth of Autoencoder, GM and DM Architectures in Different datasets}
  \label{fig:DeepKoopmanTheory}
\end{figure}

Figures 3 highlights the balance between autoencoder complexity and the effectiveness of our dataset condensation model. It suggests that increasing the autoencoder's depth enhances feature capture but also indicate that more complexity doesn't always yield better generalization. Our findings underscore the importance of optimizing autoencoder depth to produce a synthesized dataset that is both compact and representative of the original, striking a crucial balance for efficient model training.

\section{Conclusion and Future Works}

In this paper, we presented Koopcon, an innovative Autoencoder-based Dataset Condensation Model, adept at transforming large datasets into smaller, information-rich counterparts. Using an OT-theoretic criterion, we effectively secured a minimization of Wasserstein distance in Koopman space between the distributions of the original and that of condensed data. Through extensive experiments, we demonstrated that classifiers trained on these condensed datasets achieve performance comparable to those trained on the entire dataset, thus significantly reducing computational resources without compromising data integrity. Koopcon marks a pivotal advancement in data-efficient machine learning, offering feature-driven solutions in constrained environments.

Our future work will explore expansive synthesis as a complement to dataset condensation. This approach will start with a condensed but information-rich dataset and aim to expand it, generating a more diverse and informative dataset. The goal is to enhance the small dataset systematically while preserving its essential characteristics and information content. Such an expanded dataset could improve the training of machine learning models by providing a richer set of data points, potentially enhancing model robustness and generalization while managing computational costs effectively.

\vspace{12pt}
\color{red}

\end{document}